\documentclass{article} % For LaTeX2e
\usepackage[preprint]{colm2026_conference}

\usepackage{microtype}
\usepackage{hyperref}
\usepackage{url}
\usepackage{amsmath}
\usepackage{algorithm}
\usepackage{algorithmic}
\usepackage{booktabs}  % 用于三线表 toprule, midrule
\usepackage{multirow}  % 用于合并行（虽然这里主要用了multicolumn）
\usepackage{colortbl}  % 用于表格颜色
\usepackage{xcolor}    % 颜色支持
\usepackage{graphicx}  % 用于 resizebox 缩放表格

\usepackage{lineno}

\definecolor{darkblue}{rgb}{0, 0, 0.5}
\hypersetup{colorlinks=true, citecolor=darkblue, linkcolor=darkblue, urlcolor=darkblue}

\title{OPSDL: On-Policy Self-Distillation for Long-Context Language Models}

\author{Xinsen Zhang\textsuperscript{1}, Zhenkai Ding\textsuperscript{1}, Tianjun Pan\textsuperscript{1}, Run Yang\textsuperscript{1}, Chun Kang\textsuperscript{1}, \\ \textbf{Xue Xiong\textsuperscript{1}, Jingnan Gu\textsuperscript{1}}\\
\textsuperscript{1}Baidu Inc.}

\begin{document}

\ifcolmsubmission
\linenumbers
\fi

\maketitle

\begin{abstract}
Extending the effective context length of large language models (LLMs) remains a central challenge for real-world applications. 
While recent post-training methods have made progress in long-context scaling, they either rely on high-quality supervision data or sparse sequence-level rewards, leading to unstable and inefficient optimization. 
We propose \textbf{OPSDL}, an \textbf{O}n-\textbf{P}olicy \textbf{S}elf-\textbf{D}istillation method for enhancing the \textbf{L}ong-context capabilities of LLMs. 
Unlike other recent self-distillation methods that inject privileged information and rely on the model's in-context learning ability to act as a teacher, OPSDL leverages the model’s own inherently strong short-context capability as a self-teacher to supervise its own generation in long-context scenarios.
The model first generates responses conditioned on the full long-context, then the self-teacher provides per-token supervision signals via point-wise reverse KL divergence under the relevant extracted short-context. 
This dense token-level signal encourages faithful use of relevant evidence and mitigates hallucinations induced by irrelevant context. 
We evaluate OPSDL on long-context benchmarks across a range of models from 7B to 32B parameters. 
Results show consistent and substantial improvements across varying context lengths, outperforming standard post-training approaches such as SFT and DPO with higher sample efficiency. 
Notably, these gains are achieved without degrading general short-context performance. 
These findings highlight the effectiveness of OPSDL as a scalable and stable approach for long-context learning.
\end{abstract}

\section{Introduction}
Extending the effective context length of large language models (LLMs) is a central challenge in enabling real-world applications such as long-document comprehension, repository-level code understanding, and multi-hop reasoning over scattered evidence. While architectural innovations in positional encoding and attention mechanisms have made it possible to accept much longer inputs (\cite{peng2023yarn,su2024roformer,team2026minicpm}), a growing body of empirical evidence reveals a persistent gap between the \emph{maximum} context window and the \emph{effective} context capacity of a model (\cite{paulsen2025context,bai2025longbench}). These findings suggest that closing the gap between maximum and effective context length requires not only architectural advances but also innovations in post-training paradigms.

Current training approaches for long-context LLM performance optimization are numerous and effective(\cite{shen2025qwenlong,zhang-etal-2025-longreward,chen2025longpo}). 
While these methods have shown effectiveness, they suffer from either the need for high-quality training data or sparse sequence-level reward signals that limit sample efficiency. Under long contexts, such sparse, sequence-level signals make optimization extremely difficult, leading to unstable and sample-inefficient training. Furthermore, these methods typically rely on auxiliary components such as frozen verifier models or learned reward models, which increases system complexity and couples the training pipeline to the availability and quality of external resources.

To address this, we propose \textbf{OPSDL}, a simple yet effective method that overcomes these limitations. 
The key observation is that, given a query that derived from the short context, a model that given this short context naturally serves as a teacher compared to the model that given a longer contex.
OPSDL leverages this asymmetry by distilling the model’s own short-context behavior into its long-context behavior via token-level reverse KL divergence, computed on-policy during training.
Rather than relying on external supervision or reward signals, the model itself serves as both student and teacher, with the short-context ability providing a natural and readily available training signal. This design eliminates the need for human-annotated data, reward models, or elaborate reward engineering, while the on-policy distillation framework ensures that the training signal remains aligned with the model’s current behavior, yielding strong sample efficiency.

%以下待实验结论完整后，修改
We empirically validate OPSDL on the RULER(\cite{hsieh2024ruler}) across multiple backbone models and context lengths. Our results show that OPSDL consistently outperforms standard instruction tuned models on both Qwen2.5-7B family(\cite{yang2024qwen2technicalreport}), demonstrating strong generalization across model families rather than reliance on a specific architecture. Notably, the performance gains become increasingly pronounced as the context length grows, indicating that our method effectively mitigates the degradation commonly observed under long horizon settings. Moreover, despite not relying on specialized long context pretraining, OPSDL achieves performance comparable to, and in some cases exceeding, Qwen2.5-7B-Instruct-1M(\cite{yang2025qwen2}), a model explicitly trained for million-token contexts. These results suggest that replacing sparse sequence-level preference optimization with on-policy, token-level distillation enables stable policy-level self-evolution, unlocking robust long-context reasoning while preserving short-context capabilities.

In summary, our contributions are as follows:
\begin{itemize}
    \item We propose \textbf{OPSDL}, an on-policy self-distillation method that leverages a model's inherent short-context capability as a teacher to supervise its own long-context generation. By computing token-level supervision signals via point-wise reverse KL divergence, OPSDL provides dense training signals that encourage faithful use of relevant evidence and mitigate context-induced hallucinations.
    \item Unlike existing long-context training methods that rely on human-annotated data, reward models, or sparse sequence-level rewards, OPSDL eliminates the need for external supervision or auxiliary components. This design yields superior sample efficiency and training stability while remaining model-agnostic across different architectures.
    \item We conduct comprehensive evaluations on long-context benchmarks across models from 7B to 32B parameters. Results demonstrate consistent improvements across varying context lengths, with performance gains becoming increasingly pronounced as the context grows. Notably, OPSDL achieves these gains without degrading short-context performance.
\end{itemize}

\section{Related Work}
\paragraph{Post-Training for Long-Context Modeling}
A straightforward approach to improving long-context capability is supervised fine-tuning (SFT) on long-context training data, though it is constrained by data quality and prone to distribution shift. To overcome these limitations, preference optimization methods have been explored. LongReward(\cite{zhang-etal-2025-longreward}) constructs reward signals via multidimensional LLM feedback and applies DPO to boost long-context performance. LongPO(\cite{chen2025longpo}) proposes a self-evolution framework that treats short-context generations as positive samples and long-context counterparts as negative ones, iteratively optimizing the model via DPO. SoLoPO(\cite{sun2025solopo}) further decomposes alignment into short-context optimization and a short-to-long consistency constraint. 
From a training strategy perspective, QwenLong-L1(\cite{wan2025qwenlong}) treat long-context reasoning as a long-horizon decision-making problem, utilizing warm-up SFT and progressive context scaling to stabilize RL training.
However, these methods rely on externally constructed preference pairs and coarse-grained sequence-level objectives. In contrast, our approach adopts an on-policy self-distillation method with token-level fine-grained optimization signals, enabling more scalable and stable long-context learning.

\paragraph{On-Policy Distillation and Self-Distillation}
On-Policy Distillation (OPD) has recently attracted considerable attention due to its on-policy nature and token-level supervision. Pioneered by Generalized Knowledge Distillation (GKD) \citep{agarwal2024policy}, this line of work minimizes reverse KL divergence on student-generated trajectories under teacher supervision. To bypass external teacher dependency, on-policy self-distillation has emerged as a key paradigm. OPSD \citep{zhao2026self} and SDPO \citep{hubotter2026reinforcement} enable the model itself to serve as the teacher by incorporating more privileged context, such as verified reasoning traces or environmental feedback. OPCD \citep{ye2026policy} internalizes contextual knowledge and prevents forgetting by distilling from teacher-enriched contexts. While these methods typically construct a stronger teacher by enriching the input context, heavy reliance on externally provided correct trajectories may compromise the model’s intrinsic reasoning capability. In contrast, our approach takes the opposite direction: rather than enriching the context, we extract key information from long contexts to reduce noise and utilize the model’s superior short-context ability, constructing a token-level self-teacher to supervise its outputs derived from long contexts.

\section{Method}
 \begin{figure*}[t]
     \centering
     \includegraphics[width=\linewidth]{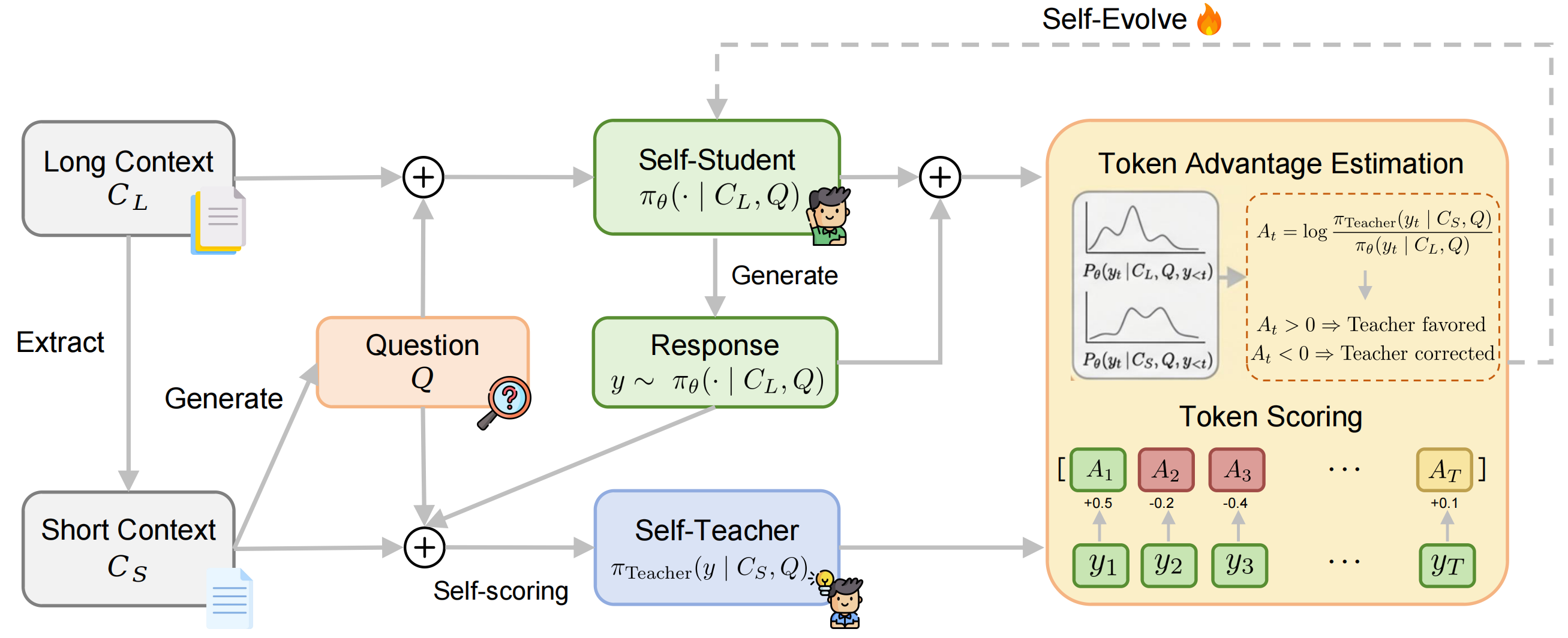}
     \caption{Overview of the OPSDL framework. Given a long context $C_L$ and its extracted short context $C_S$, the model generates responses on-policy conditioned on $C_L$. The same model under $C_S$ serves as a self-teacher, providing token-level supervision via point-wise reverse KL divergence to align the long-context generation with the short-context behavior.}
     \label{fig:framework}
 \end{figure*}

We propose \textbf{OPSDL}, which leverages the short-context capability of the current policy to supervise its own long-context generation.

\subsection{Problem Formulation}
Let $C_L$ denote the long-context input, and $C_S$ denote the corresponding short context extracted from $C_L$, which contains the core information. Based on $C_S$, an instruction or query $Q$ is formulated. Given the query $Q$ and the context, the large language model, parameterized by $\theta$ and denoted as $\pi_\theta$, auto-regressively generates a response sequence $y = \{y_1, y_2, \dots, y_T\}$. 

The key insight of our approach is that the same policy $\pi_\theta$, when prompted with the short-context input $C_S$, produces a more calibrated and accurate next-token distribution, since $C_S$ avoids distracting noise and fits perfectly within the model's well-trained context window. This allows the model to leverage its well-aligned short-context capabilities to self-align in long-context scenarios. 

Therefore, instead of relying on external reward models, we formulate the long-context optimization problem as a self-alignment process: we aim to optimize $\theta$ such that the model's generation distribution under the long context, $\pi_\theta(y \mid C_L, Q)$, closely approximates its own high-quality distribution under the short context, $\pi_\theta(y \mid C_S, Q)$.

\subsection{Data Construction}\label{sec:data_construction}

We constructs $(C_L, C_S, Q)$ triplets from raw long documents without human annotation.

We sample a long document $C_L$ from the training corpus, then extract a contiguous segment $C_S \subset C_L$ with $|C_s| \leq |C_L|$ that preserves the core evidence while fitting within the model's well-trained context window.
Finally, we generate a question $Q \sim \pi_\theta(\cdot \mid C_S)$ conditioned on $C_S$, ensuring that $Q$ targets evidence present in both contexts.
This yields triplets where the question $Q$ is answerable from both $C_L$ and $C_S$, enabling meaningful comparison between the short-context teacher and the long-context student.

\subsection{Short-to-Long Self-Distillation via Policy Gradient}

To mitigate the distraction from irrelevant information in the long context, 
we utilize the model itself as a dynamic teacher, using its predictions under 
the short context $C_S$ as a robust anchor. Formally, for a response 
$y = (y_1, \ldots, y_T)$ sampled from the long-context student policy 
$\pi_\theta(\cdot \mid C_L, Q)$, the teacher probability at each token position is:

\begin{equation}
    P(y_t) = \pi_\text{Teacher}(y_t \mid C_S, Q, y_{<t}),
\end{equation}

where the same model parameters $\theta$ are used under the short context, 
forming a self-referential teacher-student structure that evolves jointly 
during training.

\paragraph{Token-Level Advantage.}
Instead of relying on an external reward model or human annotations, we define 
a token-level advantage $A_t(y_t)$ as the log-probability ratio between the 
short-context (teacher) and long-context (student) distributions:

\begin{equation}
    A_t(y_t) = \log \frac{\pi_\text{Teacher}(y_t \mid C_S, Q, y_{<t})}
                         {\pi_{\theta}(y_t \mid C_L, Q, y_{<t})},
    \label{eq:advantage}
\end{equation}

This quantity serves as an unbiased estimator for the gradient of the 
point-wise reverse KL divergence between the two distributions, and admits 
a natural interpretation:

\begin{itemize}
    \item \textbf{Positive} $A_t(y_t) > 0$: the teacher assigns higher 
    probability than the student, indicating that the student \emph{under-weights} 
    evidence clearly present in the short context—a signal of context utilization 
    degradation under long input.
    
    \item \textbf{Negative} $A_t(y_t) < 0$: the student assigns higher 
    probability than the teacher, suggesting that the model \emph{hallucinates} 
    or attends to irrelevant content introduced by the long context.
    
    \item \textbf{Near-zero} $A_t(y_t) \approx 0$: both distributions agree, 
    indicating the token is unaffected by context length and receives negligible 
    gradient signal.
\end{itemize}

This design enables that only tokens where the long-context policy deviates from the short-context anchor receive non-trivial gradient signal, avoiding the noise introduced by uniform supervision.

\paragraph{Policy Gradient Objective.}
We optimize the model using a policy gradient objective over the long-context 
distribution $\pi_\theta(\cdot \mid C_L, Q)$:

\begin{equation}
    \mathcal{L}_{\mathrm{PG}}(\theta) 
    = -\mathbb{E}_{y \sim \pi_\theta(\cdot \mid C_L, Q)} 
    \left[ \sum_{t=1}^{T} A_t(y_t) \cdot 
    \log \pi_\theta(y_t \mid C_L, Q, y_{<t}) \right],
    \label{eq:pg}
\end{equation}

where the advantage $A_t(y_t)$ is defined in Eq.~\eqref{eq:advantage}. 
Intuitively, this objective encourages the long-context policy to increase 
the probability of tokens that the short-context teacher considers likely, 
while suppressing tokens that arise from irrelevant long-context distractions.

This reveals that OPSDL performs \emph{on-policy self-distillation} at 
the token level, where the short-context distribution acts as a dynamically 
evolving teacher signal. Unlike standard knowledge distillation with a frozen 
teacher, the teacher here co-evolves with the student, ensuring that the 
anchor remains calibrated throughout training.

\subsection{Overall Training Procedure}\label{sec:training_procedure}

Algorithm~\ref{alg:s2l_training} summarizes the complete training procedure 
of OPSDL, consisting of a \emph{data construction phase} that builds 
training triples $\langle C_L, C_S, Q \rangle$, and a \emph{training phase} 
that iteratively optimizes the policy via token-level advantage estimation.

\begin{algorithm}[H]
    \caption{OPSDL Algorithm}
    \label{alg:s2l_training}
    \begin{algorithmic}[1]
    \REQUIRE Long-document corpus $\mathcal{D}$; Initial LLM $\pi_\theta$ to be optimized; Short-context-aligned LLM $\pi_\text{Teacher}$.
    \ENSURE Optimized LLM $\pi_\theta$ with longer effective context length
    \STATE \textcolor{gray}{\textit{// Data Construction Phase}}
    \STATE Sample a long context $C_L \sim \mathcal{D}$
    \STATE Extract short chunk $C_S \subset C_L$ and generate query $Q$ via Self-QA based on $C_S$
    \STATE Construct the training triple $\langle C_L, C_S, Q \rangle$
    \STATE \textcolor{gray}{\textit{// Training Phase}}
    \FOR{each training iteration}
        \STATE \textcolor{gray}{\textit{// On-policy Rollout}}
        \STATE Generate response $y=\{y_1, y_2, \dots, y_T\} \sim \pi_\theta(\cdot \mid C_L, Q)$
        \STATE \textcolor{gray}{\textit{// Token-level Reverse-KL Advantage Estimation}}
        \FOR{each token $y_{t}$}
            \STATE Compute student log-prob: $s_{t} = \log \pi_\theta(y_{t} \mid C_L, Q, y_{<t})$
            \STATE Compute teacher log-prob: $t_{t} = \log \pi_\text{Teacher}(y_{t} \mid C_S, Q, y_{<t})$
            \STATE $A_{t} = t_{t} - s_{t}$ = $\log \frac{P_\text{teacher}}{P_\text{student}}$
        \ENDFOR
        \STATE \textcolor{gray}{\textit{// Policy Optimization}}
        \STATE Update $\pi_\theta$ using token-level advantages $\{A_{t}\}$
    \ENDFOR
    \RETURN $\pi_\theta$
    \end{algorithmic}
\end{algorithm}

\section{Experiment}
\subsection{Experiment Setup}

\paragraph{Data Construction}
Following the data preparation pipeline in LongPO(~\cite{chen2025longpo}), we construct training data from raw long documents without human annotation. We adopt \textbf{a reverse construction strategy}: for each long document $C_L$, we first randomly sample a shortened chunk $C_S$ and then prompt the model to generate instructions based on the $C_S$ via Self-Instruct(~\cite{wang2023self}). To ensure diversity, the model first generates an instruction pool and then randomly samples an instruction $I$ from it. In contrast to LongPO, our method only requires the $(C_L, C_S, I)$ triplets and does not require any preference responses. Specifically, our method trains on responses sampled on-policy from the model itself, which not only simplifies data preparation but also mitigates the distribution mismatch issues associated with off-policy data.

\begin{table}[t]
\centering
\resizebox{\textwidth}{!}{
\begin{tabular}{lcccccc|c|ccccc|c|c}
\toprule
\multirow{2}{*}{\textbf{Methods}} & \multicolumn{7}{c|}{\textbf{RULER}} & \multicolumn{6}{c|}{\textbf{LongBench V2}} & \multirow{2}{*}{\textbf{Total Avg.}} \\
\cmidrule(lr){2-8} \cmidrule(lr){9-14}
 & 4K & 8K & 16K & 32K & 64K & 128K & Avg. & Easy & Hard & Short & Medium & Long & Overall & \\
\midrule
\multicolumn{15}{l}{\textit{\textbf{7B Models}}} \\
\midrule
Qwen2.5-7B-Instruct & 94.93 & 92.87 & 92.03 & 89.20 & 68.77 & 25.14 & 77.16 & 29.2{\scriptsize±0.0} & 24.4{\scriptsize±0.0} & 33.3{\scriptsize±0.0} & 23.3{\scriptsize±0.0} & 20.4{\scriptsize±0.0} & 26.2{\scriptsize±0.0} & 51.68 \\
\quad + Long-SFT & 92.16 & 89.13 & 87.59 & 82.58 & 76.01 & 63.78 & 81.88 & 28.2{\scriptsize±0.9} & 24.9{\scriptsize±0.1} & 32.6{\scriptsize±0.7} & 22.7{\scriptsize±0.2} & 22.2{\scriptsize±0.0} & 26.1{\scriptsize±0.3} & 53.99 \\
\quad + LongPO$^\dagger$ & 95.04 & 93.33 & 89.81 & 86.40 & 74.18 & 61.28 & 83.34 & 29.2{\scriptsize±0.0} & 26.5{\scriptsize±0.6} & 38.4{\scriptsize±0.6} & 23.9{\scriptsize±0.3} & 16.7{\scriptsize±0.0} & 27.5{\scriptsize±0.3} & 55.42 \\
\rowcolor{gray!15}
\quad + Ours & \textbf{94.30} & \textbf{92.30} & \textbf{90.04} & \textbf{87.07} & \textbf{80.39} & \textbf{73.84} & \textbf{86.32} & \textbf{34.4}{\scriptsize±0.0} & \textbf{31.6}{\scriptsize±0.3} & \textbf{36.1}{\scriptsize±0.0} & \textbf{32.7}{\scriptsize±0.4} & \textbf{26.9}{\scriptsize±0.0} & \textbf{32.6}{\scriptsize±0.2} & \textbf{56.61} \\
Qwen2.5-7B-Instruct-1M & 94.28 & 93.14 & 93.39 & 90.76 & 88.15 & 81.84 & 90.26 & 36.0{\scriptsize±0.6} & 27.3{\scriptsize±0.0} & 39.3{\scriptsize±0.5} & 25.6{\scriptsize±0.0} & 26.1{\scriptsize±0.5} & 30.6{\scriptsize±0.2} & 60.43 \\
\midrule
\multicolumn{15}{l}{\textit{\textbf{14B Models}}} \\
\midrule
Qwen2.5-14B-Instruct & 96.56 & 95.26 & 93.78 & 92.20 & 77.13 & 46.69 & 83.61 & 35.0{\scriptsize±0.2} & 29.5{\scriptsize±0.3} & 37.9{\scriptsize±0.7} & 28.4{\scriptsize±0.5} & 27.6{\scriptsize±0.5} & 31.6{\scriptsize±0.2} & 57.61 \\
\quad + Long-SFT & 95.96 & 94.98 & 92.92 & 90.47 & 84.62 & 73.69 & 88.77 & 34.0{\scriptsize±0.2} & 31.5{\scriptsize±0.5} & 40.6{\scriptsize±0.5} & 28.5{\scriptsize±0.4} & 26.9{\scriptsize±1.3} & 32.5{\scriptsize±0.4} & 60.64 \\
\quad + LongPO$^\dagger$ & -- & -- & -- & -- & -- & -- & -- & -- & -- & -- & -- & -- & -- & -- \\
\rowcolor{gray!15}
\quad + Ours & \textbf{96.15} & \textbf{94.81} & \textbf{93.48} & \textbf{91.93} & \textbf{88.08} & \textbf{80.94} & \textbf{90.90} & \textbf{34.4}{\scriptsize±0.5} & \textbf{33.2}{\scriptsize±0.5} & \textbf{41.5}{\scriptsize±0.6} & \textbf{27.1}{\scriptsize±0.2} & \textbf{33.6}{\scriptsize±0.6} & \textbf{33.7}{\scriptsize±0.3} & \textbf{62.30} \\
Qwen2.5-14B-Instruct-1M & 96.81 & 96.29 & 95.46 & 94.08 & 93.01 & 89.42 & 94.18 & 38.8{\scriptsize±0.7} & 34.9{\scriptsize±0.3} & 45.6{\scriptsize±0.5} & 32.1{\scriptsize±0.7} & 29.6{\scriptsize±0.8} & 36.4{\scriptsize±0.3} & 65.29 \\
\midrule
\multicolumn{15}{l}{\textit{\textbf{32B Models}}} \\
\midrule
Qwen2.5-32B-Instruct & 96.53 & 96.23 & 95.98 & 94.09 & 81.62 & 54.02 & 86.41 & 36.2{\scriptsize±1.0} & 30.9{\scriptsize±0.8} & 39.9{\scriptsize±0.6} & 28.4{\scriptsize±0.9} & 30.4{\scriptsize±0.5} & 32.9{\scriptsize±0.6} & 59.65 \\
\quad + Long-SFT & 96.05 & 95.62 & 95.21 & 94.10 & 89.72 & 80.70 & 91.90 & 34.8{\scriptsize±0.8} & 33.9{\scriptsize±0.6} & 40.3{\scriptsize±0.7} & 30.6{\scriptsize±0.8} & 31.3{\scriptsize±0.5} & 34.2{\scriptsize±0.6} & 63.05 \\
\quad + LongPO$^\dagger$ & -- & -- & -- & -- & -- & -- & -- & -- & -- & -- & -- & -- & -- & -- \\
\rowcolor{gray!15}
\quad + Ours & \textbf{96.56} & \textbf{96.29} & \textbf{96.17} & \textbf{95.81} & \textbf{91.04} & \textbf{84.31} & \textbf{93.36} & \textbf{36.1}{\scriptsize±0.5} & \textbf{36.6}{\scriptsize±0.5} & \textbf{39.1}{\scriptsize±0.7} & \textbf{34.9}{\scriptsize±0.4} & \textbf{35.0}{\scriptsize±0.5} & \textbf{36.5}{\scriptsize±0.4} & \textbf{64.93} \\
\bottomrule
\end{tabular}
}
\vspace{2pt}
\caption{Comprehensive results on RULER and LongBench V2 benchmarks across different model sizes. For LongBench V2, each cell reports the mean over 4 independent runs at LLM sampling temperature 0.1, with $\pm$ denoting the standard deviation across runs. $^\dagger$For 7B model, LongPO results are reproduced using our curated 5K data; for 14B and 32B models, we found that LongPO failed to converge during our reproduction, and thus we do not report its results. Best results among trainable methods (excluding -1M variants) are in \textbf{bold}.}
\label{tab:comprehensive_results}
\end{table}

\paragraph{Baselines}
We compare our method with several representative long-context post-training methods. \emph{Long-SFT} performs supervised fine-tuning on long-context training data by directly minimizing the negative log-likelihood of the target responses. \emph{LongPO} minimizes the DPO loss on self-generated preference pairs, where responses produced from short contexts are preferred over those from long contexts, to reduce the performance gap between short- and long-context scenarios.

\paragraph{Models}
We adopt the \emph{Qwen2.5-Instruct} series as our backbone models, spanning three scales: \emph{Qwen2.5-7B-Instruct}, \emph{Qwen2.5-14B-Instruct}, and \emph{Qwen2.5-32B-Instruct}. For each backbone, we apply \emph{Long-SFT} and \emph{LongPO} as baseline training methods. In addition, to better assess the effectiveness of our approach in long-context settings, we compare against \emph{Qwen2.5-7B-Instruct-1M} and \emph{Qwen2.5-14B-Instruct-1M}~\citep{yang2025qwen2}, which are specifically trained to handle contexts of up to 1M tokens.

\paragraph{Evaluation benchmarks}
We evaluate our method on three representative long-context benchmarks: RULER (\cite{hsieh2024ruler}), and LongBench V2(\cite{bai2025longbench}). RULER is a synthetic long-context evaluation suite designed to quantify a model’s long-context modeling capacity. LongBench v2 is a challenging multiple-choice benchmark that evaluates deep understanding and reasoning over long documents across diverse domains and context lengths.

\subsection{Main Result}

\paragraph{OPSDL achieves the best overall performance among trainable methods.} Table \ref{tab:comprehensive_results} presents comprehensive results on RULER and LongBench V2 across three model scales. Our method consistently achieves the largest performance improvement over the base instruct model at every scale, demonstrating strong generalizability. At the 7B scale, OPSDL raises the performance from 51.68 to 56.61, surpassing both Long-SFT and LongPO. The advantages become more pronounced at larger scales.

\paragraph{Substantial improvements on long contexts.} The most striking gains appear at extended context lengths on RULER. Across all three scales, the base instruct models suffer significant performance degradation beyond 64K tokens, whereas OPSDL maintains robust performance. For example, at 128K tokens, OPSDL improves over the base instruct model by +48.70, +34.25, and +30.29 points at the 7B, 14B, and 32B scales respectively. These improvements are consistently larger than those achieved by Long-SFT and LongPO, confirming that our method, which concentrates optimization on tokens where the model's long-context behavior deviates from its short-context anchor, yields more targeted and efficient learning than uniform sequence-level supervision.

\paragraph{OPSDL narrows the gap with officially long-context-optimized models.} The Qwen2.5-Instruct-1M variants are officially released models that undergo dedicated long-context training—including multi-stage supervised fine-tuning on sequences up to 256K tokens and length extrapolation techniques—to support context lengths up to 1M tokens. These models achieve the highest absolute scores. Nevertheless, OPSDL substantially closes the gap without relying on such extensive long-context training pipelines. On RULER average, OPSDL narrows the gap to the 1M variant from 13.10 to 3.94 at 7B and from 10.57 to 3.28 at 14B. These results highlight that on-policy token-level self-distillation provides a lightweight yet effective alternative to dedicated long-context training.

\paragraph{Improvements span both synthetic and realistic benchmarks.} A key strength of OPSDL is that it improves performance on both RULER, which evaluates fundamental long-context capabilities such as retrieval and tracking, and LongBench V2, which tests realistic long-document reasoning. Across all three scales, OPSDL consistently improves LongBench V2 overall scores alongside RULER gains, confirming that our approach enhances genuine long-context reasoning rather than overfitting to synthetic patterns. Notably, LongPO failed to converge at the 14B and 32B scales during our reproduction, whereas OPSDL trains stably across all model sizes, underscoring the robustness of our training method.

\begin{table}[t]
\centering
\begin{tabular}{lcccccc}
\toprule
Methods & MMLU & ARC-C & Hellaswag & Winogrande & MT-Bench \\
\midrule
Qwen2.5-7B-Instruct    & \textbf{74.34\%} & \textbf{67.41\%} & \textbf{81.37\%} & \textbf{75.53\%} & 7.70 \\
  + Long-SFT   & 71.66\% & 63.40\% & 77.09\% & 71.51\% & 7.17 \\
  + LongPO-128K & \underline{73.40\%} & \underline{66.55\%} & \underline{80.38\%} & \underline{74.03\%} & \textbf{7.73} \\
 \rowcolor{gray!15}
  + Ours   & 73.13\% & 66.30\% & 80.01\% & 73.95\% & \underline{7.71} \\
\bottomrule
\end{tabular}
\caption{Performance on Short-Context and General Benchmarks after Long-Context Training.}
\label{tab:shortcontextresult}
\end{table}

\paragraph{OPSDL preserves short-context capabilities.} As shown in Table~\ref{tab:shortcontextresult}, OPSDL incurs only marginal degradation on short-context and general benchmarks: across MMLU, ARC-C, Hellaswag, and Winogrande, the average drop is about 1.3 percentage points relative to the base Qwen2.5-7B-Instruct model, while MT-Bench remains virtually unchanged. In contrast, Long-SFT suffers notably larger degradation (3--4 percentage points on average). These results confirm that OPSDL effectively enhances long-context performance without sacrificing the model's general capabilities.

\section{Conclusion}
We propose OPSDL, an on-policy self-distillation framework that leverages a model's own short-context capability as a dynamically co-evolving teacher to supervise its long-context generation. By computing token-level advantages via point-wise reverse KL divergence, OPSDL provides dense and targeted training signals that focus optimization on tokens where the model's long-context behavior deviates from its short-context anchor, eliminating the need for external reward models, human-annotated data, or offline preference pairs. Comprehensive experiments on RULER and LongBench V2 across the Qwen2.5-Instruct series at 7B, 14B, and 32B scales demonstrate that OPSDL consistently outperforms both Long-SFT and LongPO, with particularly striking gains at extended context lengths. Moreover, OPSDL substantially narrows the performance gap with the officially long-context-optimized Qwen2.5-Instruct-1M variants, while preserving short-context capabilities with minimal degradation. These results establish OPSDL as a simple, scalable, and effective paradigm for long-context post-training.

\bibliography{colm2026_conference}
\bibliographystyle{colm2026_conference}

% \appendix
% \section{Experiment Details}
% \subsection{Evaluation Details}
% On long-context benchmarks InfiniteBench and RULER, we evaluate our models and all baselines following the settings in the original benchmarks.  For short-context evaluation, we utilize 5-shots for MMLU.
\end{document}